# EVALUATION OF MEDIUM-LARGE LANGUAGE MODELS AT ZERO-SHOT CLOSED BOOK GENERATIVE QUESTION ANSWERING

René Peinl[1] and Johannes Wirth[2]

[1]Institute for Information Systems, Hof University of Applied Sciences, Hof, Germany
rene.peinl@hof-university.de
[2] Institute for Information Systems, Hof University of Applied Sciences, Hof, Germany
johannes.wirth.3@iisys.de

## *ABSTRACT*

*Large language models (LLMs) have garnered significant attention, but the definition of "large" lacks clarity. This paper focuses on medium-sized language models (MLMs), defined as having at least six billion parameters but less than 100 billion. The study evaluates MLMs regarding zero-shot generative question answering, which requires models to provide elaborate answers without external document retrieval. The paper introduces an own test dataset and presents results from human evaluation. Results show that combining the best answers from different MLMs yielded an overall correct answer rate of 82.7% which is better than the 60.9% of ChatGPT. The best MLM achieved 71.8% and has 33B parameters, which highlights the importance of using appropriate training data for fine-tuning rather than solely relying on the number of parameters. More fine-grained feedback should be used to further improve the quality of answers. The open source community is quickly closing the gap to the best commercial models.*

## *KEYWORDS*

*question answering, language model, survey, benchmark*

## 1. INTRODUCTION

Large language models (LLMs) are subject of many recent publications and get a lot of attention. Despite that, it is not well-defined what „large“ actually means. Whereas the BERT model with 340 Mio parameters was dubbed „large“ by its creators back in 2018 [1], in 2023 it would not be considered a LLM anymore. Later models kept the original naming convention with “small”, “base” and “large” for a while and extended it into XL and XXL to cope with the growing number of parameters. However, since the size grew even more into hundreds of billions of parameters, it became more usual to put a number suffix like 30b to designate a certain size of a model, which is more concise. Despite that, the term LLM is still used a lot in publications, and has some overlap with the term “foundation model”[2], that is defined as “[..] any model that is trained on broad data at scale and can be adapted (e.g., fine-tuned) to a wide range of downstream tasks”. In our work we refer to LLMs if the model has at least 100 billion parameters and works on text only (which excludes multimodal models). Examples for such models include Googles LaMDA [3] with 137b, OpenAIs GPT-3 [4] with 175b and Nvidia’s Megatron Turing NLG [5] with 540b parameters.

One of the benefits of such models is that they are extremely versatile multi-task models. Furthermore, their zero-shot and few-shot performance on a large number of tasks is impressive. Since they are pretrained on text-completion mostly, they are also good for “answering open-ended questions in natural language” which is e.g. explicitly mentioned in the documentation of Aleph Alpha’s Luminous.

Although the last two years were dominated by LLMs like PaLM [6] and GPT-4 [7], notable publications show that smaller models can perform nearly equally well in a lot of tasks and are much more manageable for smaller enterprises and research institutions, e.g. Chinchilla [8] with 70b parameters, FLAN T5 [9] with 11b and LLaMA [10] with up to 65b parameters. AlexaTM 20B [11] was trained for 15,360 A100 GPU days and outperforms the PaLM 540B model in 1-shot summarization (MLSum de, XSum en) and GPT-3 175B in machine translation (de-en) and SuperGLUE results. An 11B parameter model called Unicorn outperforms GPT-3 175B on CommonSenseQA 2.0 by finetuning a pre-trained T5 model on the RAINBOW datasets [12].

This paper therefore concentrates on evaluating **medium-sized language models** (MLMs) which we define as having **at least six billion parameters but less than 100 billion**. Other researchers in the meanwhile call language models still small, even if they have 11b parameters [13]. Although being quite large with 130b parameters, the GLM model should also be mentioned here, since its creators explicitly modeled it with the goal to make it accessible for researchers with less compute power [14].

Because of the increased capabilities of the LLMs, moving the evaluation to more realistic scenarios beyond purely factual answers seems necessary. The respective ML tasks are called long-form question answering [15] which was originally designed to involve document retrieval before answering the question. However, LLMs and to some degree also MLMs should be capable of performing it as closed-book QA [16]. They pose the problem that evaluation of results is difficult due to ambiguity of questions (ibid) and other challenges. Due to that, answers of models are hard to evaluate with wide-spread methods like ROUGE [17]. We therefore perform a human evaluation to test model accuracy. The remainder of this paper is structured as follows. We first discuss related work, especially other evaluations of language models. Then, we introduce the dataset used for the evaluation before discussing the choice of models for the test. After that, the experimental setup is described, before AI results and a human baseline are outlined. The paper ends with limitations, conclusion and outlook.

## 2. Related Work

To evaluate the performance of LLMs a plethora of benchmarks and datasets were published, e.g. Natural Questions [18], BIGbench [19] and MMLU [20]. However, they all concentrate on questions that are automatically evaluable, which means they either test using multiple-choice questions, which limits the language generation capabilities of the LLMs to generate a single character, or use measures like ROUGE and BLEU, who are known to have severe limitations regarding their ability to identify correct answers that deviate from the wording of the ground truth [21].

Results from holistic evaluation of language models (HELM) [22] confirm the assumption of this paper, that well-tuned MLMs can outperform much larger models. The 52b parameter model from Coherence (v20220609) outperforms the 175b parameters models like GPT-3 davinci v1, J1-jumbo v1 and Bloom. The 52b parameter model from Anthropic (v4-s3) performs even better and additionally outperforms OPT-175b and nearly reaches the accuracy of Turing NLG v2 with 530b parameters. However, the performance of MLMs is very different based on their training, esp. fine-tuning data. T0++ for example performs third best in TruthfulQA EM and outperforms all larger models except GPT-3 175B, whereas it is beaten by much smaller models like GPT-J 6B and all of the bigger ones on NarrativeQA closed-book F1. Similary, UL2 20B performs relatively well in NaturalQuestions closed-book F1 nearly reaching the performance of Bloom 175B, but is beaten by GPT-J 6B and Bloom 175B in Truthful QA EM (although only by a small percentage).

Another important outcome from [22] is, that “automated evaluation was not satisfying” and it is “necessary to conduct human evaluations to better understand language model performance”. This is done in our paper. The goal of this paper is further similar to HELM in the intend to use the same benchmarks on all considered models instead of a sparse evaluation matrix of tests and models. Besides that, the evaluation in this paper has only a small overlap in models considered (e.g., T0pp and T5) and proposes own tests instead of reusing the popular ones. The evaluation of LLMs in [2] does a good job of summarizing developments of the last years but is not concerned with own benchmark results. Mahowald et al. [23] analyze LLMs from a linguistic perspective and differentiate between formal and functional linguistic competences. Based on literature analysis they reach the conclusion that LLMs are highly competent although not perfect in formal linguistic competence but often fail on functional linguistic competence. The examples they state are however kind of artificial ("How to get a sofa onto the roof of a house") and also overcome by newer models like LamDA or ChatGPT (like the trick question to translate a sentence that includes a new direction), which they hint to in stating that they are only talking about models trained without human reinforcement or instruction tuning (p. 9).

One major improvement in the advancement of LLMs is using instruction tuning [24]. U-PaLM [25] significantly increases zero-shot performance of PaLM with only 0.1% extra compute, by applying the mixture of denoising training objective from UL2 [26] to a pretrained PaLM model. Flan-PaLM [27] further improves on that by using both instruction-tuning and chain-of-thought prompting. The relative improvement is even greater for the 11B parameter T5 XXL model (+26.6%) compared to the 540B parameter PaLM model (+9.3%).

Similarly, Suzgun et al. [28] find that chain-of-thought (CoT) prompting dramatically increases the accuracy of LLMs in hard BIGbench tasks. PaLM, InstructGPT and Codex benefit with at least 12.9% absolute accuracy increase from low 50ies to high 60ies. The highest increase was found for Codex in the algorithmic tasks (+28.5%). However, for smaller model sizes (8B) there was a negative impact using CoT. For extremely hard tasks, CoT prompting helped the model to create emergent capabilities although those tasks seemed to be not affected by model scale and would require complete new architectures. [29] collect a large number of instructions in order to finetune MLMs on diverse -tasks and achieve good results. Similarly, [30] perform finetuning but use an automatically generated dataset to achieve comparable accuracy on the BIG-bench hard subset.

Multi-step reasoning is still challenging for LLMs [31]. One example for advancement in this area is the Self-Taught Reasoner (STaR) introduced by [32], in which a LLM is trained and refined on its own output iteratively. Specifically, with CoT prompting, the model first generates initial rationales. And then, the model is finetuned on rationales that lead to correct answers. As a follow-up to this work [33] show that LLMs are able to self-improve their reasoning abilities without the need for supervised data by leveraging the self-consistency of reasoning. Benchmarks that can be used for testing commonsense reasoning [31] abilities of LLMs include CSQA, StrategyQA and ARC. We refer the reader to Bhargava and Ng (2022)'s survey for more work in this domain. According to [34], LLMs exhibit reasoning patterns similar to those of humans as described in the cognitive literature.

## 3. DATASET

Our own dataset is self-constructed and takes some inspiration from existing datasets like BigBench, TriviaQA and AmbigQA. The following categories are included.

- **Abstractions** replace one well-known concept with a different one and force the model to answer based on the replacement.
  Example: Assume that purple represents a car and red represents a roof. What do you get if you remove the red part from purple?
- **Basic physics** requires some background knowledge and its application to more or less common situations.
  Example: If a ball drops from 2 meters height onto the floor and the floor is made of stone and the ball is made of glass. What happens to the ball?
- **Everyday knowledge** is easy for humans to answer, but unlikely to be found 1:1 in the training data.
  Example: 10 year old John is going shopping with his grandfather Raymond. Who is more likely to want to buy some cigarettes?

- **Trick questions** are made to fool humans and it is interesting to see whether the AI can be fooled in the same way.
  Example: Which weighs more, a pound of silver or a pound of gold?
- **Metaphors** use well-known English sayings or phrases and turn them into a question. It requires recognition of the saying that is presented in a slightly different form and an understanding of the metaphoric meaning
  Example: What kind of coals do you need to take coals to Newcastle?
- **Math word puzzles** are known to cause problems for LLMs. We therefore only include a few of them and also combine them with questions that look mathematical but need no calculation for a correct answer.
  Example: If Susan is running faster than Joe, but slower than Mike and the three do a 100 meter race, who will win?
- **Relational reasoning** transfers the rule of three to everyday objects and requires to understand similarities and differences.
  Example: A house relates to a skyscraper like a flower relates to what?
- **Deductive reasoning** requires to derive conclusions from the premises of the question. Example: If the flow of time causes the hands of a clock to turn to the right, what happens if time could run backwards?
- **Symbolic reasoning** is a bit similar to abstractions but uses short variable names instead of words that are defined in a different way as replacements.
  Example: If x is a boy and X is a man, what is y if Y is a woman?

Often, a (missing) deeper understanding of the model can be seen when comparing the answers to related questions. In the basic physics category, there are several questions regarding balls dropping on the floor and only the height or the ball material is varied. If the answers reflect this variation, the model seems to be able to capture the required understanding. If it always answers “it bounces” no matter whether the ball is made of rubber, steel or glass, it shows that the model did not understand. We also did vary the wording to find out if it makes a difference how questions are asked. The dataset will be published on opendata.iisys.de.

## 3. CHOICE OF MODELS TESTED

The primary source for models to be tested was huggingface. Models were included if they fall in the medium-size category, are pretrained at least in English language (multi-lingual models were included as well) and preferably already finetuned on closed-book question answering or instruction-tuned in general. However, we also included models without any finetuning. Models that were trained for extractive question answering instead of generative were excluded as well as those that need a document retriever. Models that are not publicly available like PaLM 62B [6] or Chinchilla 70B [8] were excluded as well.

At the beginning of the study in November 2022, the only model available in multiple sizes of the MLM type was OPT [35] from Meta. In order to study scaling effects, all four models from 6.7b to 66b parameters were included. With OPT-IML and Galactica there are also two 30B parameter variants available that build on OPT-30B and add instruction tuning. Later on, Meta released LLaMA [10] and soon after that Stanford published the instruction-tuned LLaMa version Alpaca [36]. Following these two releases, a number of derivatives and similar models have been published including Vicuna [37], [38] and Databricks’ Dolly [39]. Whereas Alpaca and Vicuna are based on LLaMA, Dolly v1 is based on GPT-J 6B and v2 on Pythia, an open model from Eleuther AI [40]. Furthermore, several versions of T5 [41] were added to study the effect of different finetuning methods and datasets. These include Flan-T5 [27], mT0 [42], T0pp [43] and T5-SSM-TQAO [41]. ChatGPT from OpenAI, which is a fine-tuned version of GPT-3.5 with 175B parameters and the largest GLM model with 130b parameters serve as a reference for truly LLMs.
The goal was to include as many and as recent models as possible, so models from the collectives BigScience [42], [44] and Eleuther AI [45] have been included as well as models from the Allen AI institute [29], Stability AI and Bejing AI [46]. So BloomZ with its 7b parameters can be fairly compared to OPT 6.7b, GPT-J 6b and GPT-JT 6B. GLM 10b can be compared to the T5 variants with 11B parameters and LaMA 13b. GPT Neo-X with its 20b parameters is a bit in between and should be compared to both the 13b and 30b models.
Models that were explicitly geared towards dialog like Guanaco, HuggingChat, Koala and OpenAssistant[1] were not included in the comparison and are planned for a future analysis with special focus on chatbots.
Late additions in the test were 4 bit models provided by Huggingface user TheBloke that used finetuning in 4bit and therefore allowed much larger models to be finetuned with limited resources like WizardLM 30B and Wizard Vicuna 30B, as well as models published beginning of June 2023 like Luminous Supreme Control, Falcon 40B Instruct and Dromedary 65B. No models with experimental 8k or larger context size were included.

## 3. Experiments

We followed the instructions of the creators of the MLMs, e.g. by using prefixes like “q: “ before and “a: “ after the question, or “please answer the following question:” as instruction. We did not use any prompt-engineering or chain-of-thought prompting. Except the LLM references ChatGPT and Luminous which were used as part of their manufacturers’ cloud offerings, all models were run on an A100 80 GB GPU (or multiple if necessary) on our local server with FP16 precision. To make results more reproducible we set the temperature value to 0.1.
Open-ended questions have the problem, that they cannot be easily evaluated in an automated way. It is not only possible to give the correct answer in an alternative formulation that might not be detected by current evaluation methods like BLEU and ROUGE [47], but there were also answers given by the language models that were correct and surprising to humans so that even advanced methods like BERTscore [48] would not help detecting the correctness. Flan-Alpaca for instance answered “Tempura” to the question “What relates to Japan like pizza relates to Italy?“. The ground truth answer was “Sushi”, but Tempura seems an even better answer since it is also well-known and additionally closer related to pizza than Sushi. Therefore, a manual evaluation of the answers was performed. Initially, the answers were rated per model. Later on, a cross-check per question across models was performed to assure an equal treatment of each model, since human evaluation comes with the risk of subjectivity.

## 3. AI Results

In initial tests, **BloomZ** was the best model in the 7B parameter range with 35.5% accuracy. It outperforms Alpaca 7B (chavinlo, 33.6%) in our experiments, but only by a small margin (see table 1). Alpaca is based on LLaMA 7B and chavinlos model is not improving the already good base performance of LLaMA a lot (32.7%). However, it was unclear whether the replication of Stanford’s Alpaca that is hosted publicly on Huggingface (chavinlo/alpaca-native) is really performing as good as the original. The keyword alpaca produced 605 results on huggingface (2$^{nd}$ of May 2023). Most have no model-card and several did not run with the code we used for testing LLaMA. However, using Wenxiang Jiao’s **Alpaca** 7B repository produced the surprising result that it was performing not only much better than the first Alpaca model tested, but outperformed all other MLMs with 7b parameters up to this date with 46.4% correct answers. How much of an improvement instruction tuning can give is also visible for **GPT-J** and its fine-tuned version GPT-

1 https://huggingface.co/OpenAssistant/oasst-sft-6-llama-30b-xor
https://huggingface.co/KBlueLeaf/guanaco-7B-leh
https://huggingface.co/TheBloke/koala-13B-HF

JT. The latter improves the rather bad 17.3% performance of the base model to 28.2%. This is however still worse than the **Dolly** v1 version with 6B parameters, which is also based on GPT-J and scored 31.8%. Instruct GPT-J further pushes this score to 39.1%. Surprisingly, Dolly v2 does not score better than v1 but only 30.0%, although it has twice the number of parameters. Its base model **Pythia** 12B scores 19.1%, which is also worse than expected.

Table 1. Results of models with 7B parameters and less (33.6% avg.)

| Name | Accuracy | Name | Accuracy |
|---|---|---|---|
| Alpaca (chavinlo) | 33.6% | LLaMA 7B | 32.7% |
| Alpaca (wxjiao) | 46.4% | MPT-7B-Instruct | 40.9% |
| BloomZ 7B | 37.3% | OpenLLaMA 7B Instruct | 31.8% |
| Dolly v1 6B | 31.8% | OpenLLaMA 7B OpenInst. | 30.0% |
| Falcon 7B Instruct | 40.0% | OPT 6.7b | 18.2% |
| GPT-J 6b | 18.2% | StableLM 7B | 11.8% |
| GPT-JT 6B | 28.2% | **WizardLM 7B** | **47.3%** |
| Instruct GPT-J 6B | 39.1% | Wombat 7B | 44.5% |
| | | Wombat 7B GPT4 | 40.9% |

**StableLM** performed exceptionally bad for an instruction tuned model with 11.8% correct answers and also stood out from the rest by often stating "As an AI language model, I do not have personal beliefs or opinions." It also suspected ethical problems where none were visible e.g., the question "A ten-year-old girl and a 30 year old man sit together in a restaurant. Who is more likely to work there?".

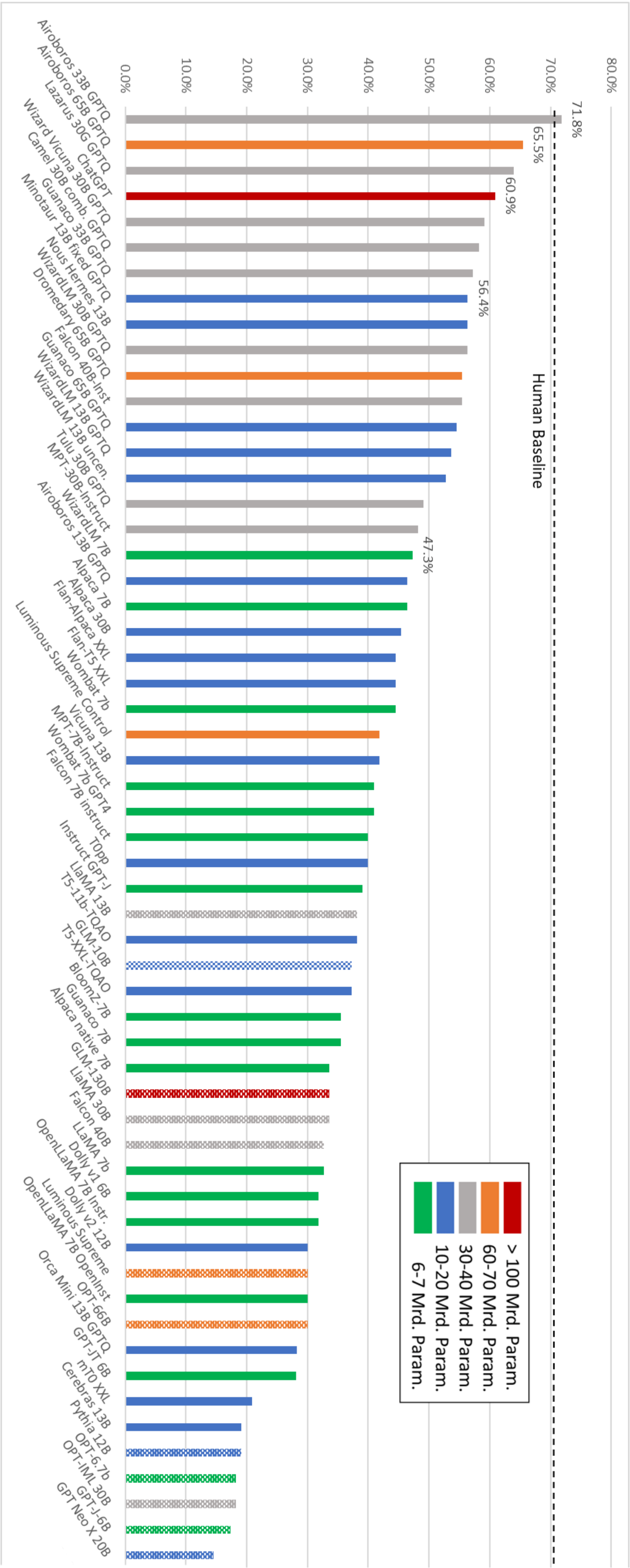

Figure 1. Performance of the best performing models compared to the human baseline

A last minute addition was **Wombat**, another finetuned LLaMA version, but this time with a reinforcement learning approach [49]. It is available in two versions with instructions generated by ChatGPT and GPT-4 respectively. Both perform very good and nearly reach Alpaca's performance. However, they behave quite differently since the GPT-4 instructed model gives quite concise answers (either right or wrong), whereas the other version produces very verbose answers and starts nearly every answer with "As an AI language model, I do not have personal beliefs or opinions". With Falcon 7B and MPT-7B, two strong competitors joined the field in June 2023 with 40% and 40.9% correct answers. They both rely on own pretrained models and can therefore be used commercially, in contrast to the LLaMA-based alternatives. Finally, WizardLM took the lead in the 7B parameter models with 47.3% correct answers.

The finetuned **T5** family of models performed rather good in our tests (see table 2). Scores reach from 37.3% to 44.5% of Flan-T5. However, they still perform slightly worse than the smaller models Alpaca 7B and WizardLM 7B. Flan-Alpaca and Vicuna 13B perform similarly good. T5 also shows how much of an effect finetuning has since the base model scores only 13.6% which means 24% to 30% increase. One notable exception is **mT0** xP3 XXL, which is a multi-lingual version of T5. It seems that its subpar performance with only 20.9% correct answers is due to the multi-lingual pretraining since BloomZ 7B is also finetuned with xP3 and shows very good results.

**Vicuna and Flan-Alpaca** tend to give correct, but longer answers on average (93 and 115 characters) compared to the T5 models (~16 characters) and also compared to LLaMA, GLM and BloomZ (between 50 and 15 characters). They are more similar to ChatGPT in this respect (128 characters). **WizardLM** [50] performs very well in 4 bit GPTQ version with 13b parameters from TheBloke (53.6%) and therefore clearly outperforms Flan T5. The direct comparison shows, that the quantization indeed does not decrease performance, since the 4bit model performs even slightly better than the FP16 model. However, WizardLM is beaten in the 13B param category by Nous Hermes and Minotaur, that both are very close to ChatGPT's performance with 56.4%.

**Orca**-Mini is a first try to mimic the training strategy of Microsoft's Orca [51]. However, it is failing miserably and achieves only 28.2% correct answers whereas the original Orca is able to outperform ChatGPT in most tasks and gets even close to GPT-4's performance.

Table 2. Results of models with 10B to 13B parameters (35.4% avg.)

| Name | Accuracy | Name | Accuracy |
|---|---|---|---|
| Airoboros 13B 4b | 46.4% | Cerebras 13B | 19.1% |
| Dolly v2 12B | 30.0% | Flan T5 XXL | 44.5% |
| Flan Alpaca XXL | 44.5% | GLM 10B | 37.3% |
| GPT Neo X | 14.5% | LLaMA 13B | 38.2% |
| mT0 xP3 | 20.9% | **Minotaur 13B fixed** | **56.4%** |
| **Nous Hermes 13B** | **56.4%** | OPT 13B | 10.9% |
| Orca Mini 13B 4b | 28.2% | Pythia 12B deduped | 19.1% |
| T5 v1.1 XXL | 13.6% | T5 XXL SSM TQAO | 37.3% |
| Vicuna | 41.8% | T5-11b-TQAO | 38.2% |
| WizardLM 13B | 52.7% | WizardLM 13B 4b | 53.6% |

The **OPT** family of models showed subpar performance with 11% correct answers for the 30B parameter model (see table 3) and seems to prove the warning you often read, that pre-trained models without any finetuning are not usable for downstream tasks. However, if you consider **GLM**-10B and **LLaMA**-13B, the models achieve 37.3% and 38.2% correct answers without any finetuning, the statement doesn't seem to be correct. They are therefore in the same performance range as T5 finetuned on closed book QA and instruction tuned T0pp with accuracies between 37.3% and 40%. The larger models OPT-30B and LLaMA-30B did not outperform their smaller siblings. OPT-66B was significantly better than the smaller OPT models, but still substandard given its size. OPT-IML 30B and **Galactica** 30B with 18.2% and 12.7% respectively, were also rather disappointing. We could not produce usable results with LLaMA 65B, which may point to an erroneous checkpoint being leaked / published.

The 70B parameter model Luminous Supreme from the German startup Aleph Alpha performs similar to OPT-66B with 30% accuracy. In June, a new version of it called "instruct" was published that increased this result to 41.8%, which is still below the best 7B parameter models. After the publishing of QLora in May 2023 [52], a lot of new instruction-tuned models with 30B parameters and more were published. This pushed the previously low average score of the largest models significantly. However, the gain for well-trained smaller models is not large. WizardLM's performance increases from 53% to 56.4% by increasing the model size from 13B to 30B parameters. MPT jumps from 40.9% (7B) to 48.2% (30B) accuracy. Alpaca even drops from 46.4% (7B) to 45.5% (30B). However, a few other competitive models got available. Falcon 40B instruct achieved 55.5% accuracy, the same as IBM's **Dromedary** with 65B parameters and GPTQ [53]. Allen AI's **Tulu** 30B performed only mediocre with 49.1% compared to Caldera AI's **Lazarus** 30B that achieved 64% and therefore outperformed ChatGPT (60.9%). The best overall models however came from a single developer called Jon Durbin. His **Airoboros** model outperforms ChatGPT and Lazarus with 65.5% accuracy (65B 4bit model) and even outperforms the human baseline with 71.8% with its 33B parameter model.

Table 3. Results of models with 30B parameters and more (44.8% avg w/o ChatGPT)

| Name | Accuracy | Name | Accuracy |
|---|---|---|---|
| OPT 30B | 11.8% | LLaMA 30B | 33.6% |
| OPT 66B | 30.0% | LLaMA 65B | 0%[2] |
| OPT-IML 30B | 18.2% | GLM-130B | 33.6% |
| Alpaca 30B 4b | 45.5% | Camel 30B comb 4b | 58.2% |
| Dromedary 65B 4b | 55.5% | Falcon 40B instruct | 55.5% |
| Galactica 30B | 12.7% | MPT-30B instruct | 48.2% |
| Luminous Supreme | 30.0% | **Lazarus 30B 4b** | **64.0%** |
| Luminous Supreme Control | 41.8% | **Airoboros 33B 4b** | **71.8%** |
| Tulu 30B 4b | 49.1% | **Airoboros 65B 4b** | **65.5%** |
| WizardLM 30B 4b | 56.4% | **ChatGPT 3.5** | **60.9%** |
| Wizard Vicuna 30B 4b | 59.1% | *Human baseline* | *70.1%* |

## 3. Human Baseline

A test with different groups of humans was performed to determine a human baseline per category of questions. This is not only used to compare the performance of MLMs but also to verify the judgement of what a correct or plausible answer should look like. All participants were non-native English speakers but had a good English speaking level so that they can attend English study programs. 32% had a background in computer science, 22% in business administration, another 22% in engineering and the final 22% other. The questions that were asked to the AI models were split into four questionnaires, so that every participant saw only parts of the whole question catalogue which helped keeping the time to answer within bounds (15.5 min median). The participants did not get any incentives for participating.

2 We tested the model with a number of different prompts and hyperparameters, but it kept on repeating the question instead of giving answers. We also tested different models on huggingface, but all had the same issue.

Overall, there were 87 participants (41.4% female, 54.0% male) who finished the questionnaire (dropout 8.4%). The median age was 23 years (avg. 25.8).
It was expected that humans would in general be able to answer the questions, but would also make a couple of mistakes, so that the baseline would be around 90%. Astonishingly, the average human score was only 70.1%. Questions were partly trivial to answer for humans, but also partly challenging. As expected, a significant portion of humans had problems both in math word questions as well as abstractions and symbolic reasoning. They also fell for some of the trick questions. Some had problems with missing background knowledge especially for historic celebrities like Margaret Thatcher or Edwin Moses due to their young age. Still, some of the subjectively trivial questions like relations between animal types resulted in surprising answers (e.g. donkey and zebra as an answer to "A tiger relates to a wildcat like a horse relates to what?" instead of pony). On the other hand, some of the questions that were rather controversial because they seem underspecified, were answered relatively homogeneously in the expected way. Especially age-related behavior was only scarcely questioned and over 95% of participants agreed that children are more likely to eat ice cream than their grandparents and vice versa for cigarettes. It can be seen as confirmation of prejudice or as a Fermi question, but most respondents agree that females are more likely to buy hair color than males (87%). Regarding the trick questions there were large differences. Some of the obvious ones were answered worse than expected, e.g. only 26% of respondents noticed, that an electric train does not produce smoke and only 5% found that the car which can drive up to 120 km/h won't accelerate to 200 km/h. On the other hand, only 25% fell for the question about how long bamboo needs to grow to 30m height, if it can grow up to 20m tall, although it is quite similar.
Surprisingly, the wrong answers from humans were the same or very similar to those of the AI models, even if the question did not push them into the wrong direction as it is the case in the trick question category. Both humans and AI e.g., saw the symbols XYxx as an indication of transgender instead of a family with two boys since it was defined in the question that X represents a man and Y a woman.

## 3. DISCUSSION

On average, the MLMs were able to answer 35.3% of questions correctly, which means just over a third. We therefore conclude that our dataset is challenging. The best performing model was Airoboros 33B which scored 71.8% and therefore clearly above the reference LLM ChatGPT (60.9% correct answers) and even slightly above the human baseline. Some of the larger MLMs, especially the 30B parameter models were somewhat disappointing since they did not outperform their smaller siblings. However, the best models were still from this category. It is also surprising that the correct answers of the models, especially Flan-T5 and Vicuna 13B are somewhat complementary. The 44.5% and 41.8% of the models add up to 62.7% correct answers which even outperforms ChatGPT (60.9%). You would expect that questions are either harder or easier for models to answer and that well performing models give good answers to the same questions, if the questions were not included in their training data. However, across all MLMs the correct answer rate was 91.8% and together with ChatGPT only 4 questions could not be answered correctly.
**Short answers** are preferable for factual information, while longer answers are suitable for fact-based judgments. The models allow for additional parameters, such as specifying the maximum number of tokens for the answer. However, this often results in truncated answers that abruptly end in the middle of a sentence, rather than providing shorter responses. Ideally, the model should autonomously distinguish between answers that are better when concise and those that require additional explanations. However, there is also a subjective notion to that judgment. An MLM's poor quality is evident when it generates unrelated text or merely reproduces training data without providing a relevant answer. A similarly bad behavior is **asking new questions** that are almost identical to the original question but do not contribute to a proper answer. It sometimes seemed, as if this was the MLMs way of saying: “I have no idea”. The ability to **confess being unknowledgeable** is lacking in all models but GPT Neo X.

**Filtering** to avoid biases seems rather undesirable. It would be better to train the model for desirable answers. Not only ChatGPT with filtering, but also StableLM and WizardLM without any filtering showed signs of trying to teach the user, e.g. in preaching healthy living styles without smoking when being asked about wildfires that are caused by smokers. This seems also undesirable, although in general giving advice for self-improvement of the user is good. Another similar issue is the answer: “It is not appropriate to make assumptions about a person's personal preferences based on their age.” given by StableLM on the question about the likelihood of buying something based on age.

**Repeating the question as part of the answer** has pros and cons but is rather undesirable. Luminous has even an option for penalizing this, although it did not seem necessary there. ChatGPT on the other hand does that very frequently.

**Generating options before giving an answer** must be viewed as an undesirable feature and is present in many models that are not finetuned. Galactica is one of the worst regarding that. Sometimes models do generate only options with no choice afterwards, or the options were so long, that they did not fit into the maximum answer length. This was rated equally to unanswered or wrong answer.

**Hallucination** is a well-known problem of LLMs and not surprisingly, the problem was observed for MLMs as well. To go into more detail, a conspicuous situation that seems to force MLMs to hallucinate are questions regarding similarities. This is a situation where humans as well might get into speculating if they do not find an obvious similarity. Since LMs in general do have problems in confessing that they do not know about certain things, it is not surprising, that they invent similarities between the celebrities, e.g. common birthplace, age of dying, art or sport area and so on.

**Prompt engineering** should only be an intermediate step towards better language models, since it should not be the task of a human to ask the question in a way that allows the LLM to give the correct answer, but the LLM should be trained in a way that allows it to understand all kinds of questions and always gives the best possible answer (given its training data). For Luminous for example, it made a great difference whether it is prompted with the context and question only, or there was a prefix “question: “ before the actual question. It did not help to put the “question: “ prefix before the context. With the prefix, the answers were much better than without. It is even very picky regarding some wordings, e.g. it is more likely to produce correct answers if you start the context with “let’s assume” instead of just “assume”.

**Mathematical capabilities** of the models tested are very different. Some models are able to perform some basic calculations like adding, subtracting, multiplying and dividing and use these capabilities to solve some simple math word problems. However, in most cases they struggle if there are too many calculations involved, even if they are simple to calculate. They also mostly fail to do unit conversions, e.g. from meters to centimeters or from kilometers per hour to meters per second. Astonishingly, ChatGPT is able to do the latter, but unable to correctly perform the former, although it recognizes that it has to do a conversion.

Regarding **scale**, there was no clear tendency. Although larger models performed better in general (e.g. LLaMA 13b was 5.5% absolute better than LLaMA 7b), both LLaMA and OPT 30b models performed worse than the smaller models. Also, Galactica 30b and OPT-IML 30b were not as good as expected and even the 70B Luminous Supreme model performed worse than several 13b and even 7b parameter models. The assumption is, that the 30B models are **undertrained** compared to the smaller models. This finding is in line with the degradation of LLaMA 65B compared to LLaMA 33B in zero-shot settings for NaturalQuestions, ARC-e and ARC-c [10]. Typically, larger models with enough training outperform smaller models in every aspect and especially in zero-shot performance. We also hypothesize that instruction-tuned models perform better the **more compute was invested in their finetuning**. Another reason could be that for demanding tasks a low score of around 20% accuracy is still in the area where chance plays a role. The empirically observed hockey-stick curves when scaling language models and evaluating their performance compared to scale seems still in the "blade" area of the curve and not yet in the "shaft" area.
Comparing the performance of OPT and LLaMA, the latter models perform way better than OPT, so there is an a**dvancement from OPT over OPT-IML to LLaMA**. This seems to be due to increased training data and also epochs of training. Meta doesn't state exactly how long the OPT models have been trained, but the usage of 992 A100 GPUs compared to the 2048 for LLaMA together with the increase in training tokens from 180 B (OPT) to 1.4 T for LLaMA suggest that OPT is heavily undertrained and LLaMA compares to OPT similar as Chinchilla [8] compares to Gopher.

## 3. LIMITATIONS

Inference time was not measured explicitly, but never exceeded a few seconds (<5). per question on an A100, depending on the number of tokens produced (<100) and the model size (<30B). The human evaluation was done by the authors only. For future work, there should be a cross-check with automatic evaluation based on GPT-4 as proposed in [38] and more human evaluators should be included.
Only a small number of test questions was used (110 altogether). This kept the effort for human evaluation within bounds, but as a downside, tests only a limited amount of application areas, e.g., no questions were included that tested for racial or gender bias explicitly.
The human baseline was limited to students and university staff and all were non-native speakers of English language. We did not perform any further analysis of correlations between number of correct answers in a specific question category and the academic background of the participants yet.
The evaluation was done with FP16 for all models, initially. Later on, a few 4bit quantized models were tested using the GPTQ for Lora framework. We tested a few models in both FP16 and int4 and could not find a significant difference. Therefore, models available in GPTQ format were used wherever possible since June 2023.
The bad performance of several larger models with 30B and more parameters is astonishing, and it cannot be completely excluded that there are technical problems involved when using multiple GPUs for inferencing. We ran the models based on the advice in the corresponding papers and model cards to the best of our knowledge, but independent verification of the results is necessary.

## 3. CONCLUSIONS AND OUTLOOK

If you take together all the right answers from the different MLMs, 91.8% of the questions were answered correctly. The question remaining is therefore, how to combine the best of all models into a single model within a range of 7-30B parameters. It seems that using the **right training data for finetuning** is more important than the pure number of parameters. However, this finding might be due to a similarity of some training data to our own dataset. It was beyond the scope of this paper to make a detailed evaluation of the overlap between questions in our test dataset and the training data of each model tested. We assume that the overlap is rather small, since we took quite some efforts to come up with unique questions. Only the trick questions are likely to be included in training data, since they were taken from the internet. However, performance on those was rather bad. Only one model correctly answered the question about getting out of an imaginary room and none was able to figure out, an electric train does not produce smoke.
**Instruction-tuning and RLHF** provide a much better training resulting in models giving substantially better answers than models without this kind of finetuning as stated in literature [27], [36], [50], [54]. However, to unlock the full potential of MLMs, they would need even more fine-grained feedback. The longer the answers get, the less an aggregated score summarizing the human preference for the answer helps. Promising future directions are to analyze ways to target the feedback to certain parts of the answer. Consider multi-hop reasoning for example. Was the first step already faulty, or was it the final conclusion that did not fit to the previous intermediate results, although these were correct? This makes a big difference and would be problematic for humans as well, if we would always just get aggregated feedback. Imagine writing a two-page essay and getting the grade as the only indicator on how well you performed. It would be very hard to get better at writing essays with this kind of feedback and no alternatives for learning. We need a similar development as it was performed for sentiment analysis when moving from an overall rating of a product review (positive/negative) to aspect-based sentiment analysis [55], that provides much finer grained judgements that are much more helpful. During the review process of this paper, OpenAI published own findings that support this claim [56].

## ACKNOWLEDGEMENTS

The authors would like to thank everyone, just everyone!

## APPENDIX A

Table 4. Alphabetical list of models used for the evaluation with size in billion parameters.

| Creator | Model | Based on | Size [b] | Training |
|---|---|---|---|---|
| Jon Durbin / TheBloke | Airoboros 33B GPTQ | LLaMA | 33 | Inst |
| Jon Durbin / TheBloke | Airoboros 65B GPTQ | LLaMA | 65 | Inst |
| Caldera AI | Lazarus 30G GPTQ | LLaMA | 30 | Inst |
| OpenAI | ChatGPT | GPT 3.5 | 175 | RLHF |
| Eric Hartford / TheBloke | Wizard Vicuna 30B GPTQ | LLaMA | 30 | Inst |
| Camel AI | Camel 30B comb. GPTQ | LLaMA | 30 | Inst |
| OpenAccess AI Collective | Minotaur 13B fixed GPTQ | LLaMA | 13 | Inst |
| Nous Research | Nous Hermes 13B | LLaMA | 13 | Inst |
| Microsoft / TheBloke | WizardLM 30B GPTQ | LLaMA | 30 | Inst |
| IBM | Dromedary 65B GPTQ | LLaMA | 65 | Inst |
| Technology Innovation Institute, UAE | Falcon 40B-Inst | Falcon | 40 | Inst |
| Microsoft / TheBloke | WizardLM 13B GPTQ | LLaMA | 13 | Inst |
| Eric Hartford | WizardLM 13B uncen. | LLaMA | 13 | Inst |
| AllenAI | Tulu 30B GPTQ | LLaMA | 30 | Inst |
| MosaicML | MPT-30B-Instruct | MPT | 30 | Inst |
| Microsoft | WizardLM 7B | LLaMA | 7 | Inst |
| Jon Durbin / TheBloke | Airoboros 13B GPTQ | LLaMA | 13 | Inst |
| Stanford / wxjiao | Alpaca 7B | LLaMA | 7 | Inst |
| Stanford / chansung park | Alpaca 30B | LLaMA | 30 | Inst |
| DeCLaRe Lab | Flan-Alpaca XXL | T5 | 11 | Inst |
| Google | Flan-T5 XXL | T5 | 11 | Inst |
| Alibaba | Wombat 7b | LLaMA | 7 | RRHF |
| Aleph Alpha | Luminous Supreme Control | Luminous | 70 | Inst |
| lmsys.org | Vicuna 13B | LLaMA | 13 | Inst |
| MosaicML | MPT-7B-Instruct | MPT | 7 | Inst |
| Alibaba | Wombat 7b GPT4 | LLaMA | 7 | RRHF |
| Technology Innovation Institute, UAE | Falcon 7B instruct | Falcon | 7 | Inst |
| BigScience | T0pp | T5 | 11 | Inst |
| Crumb | Instruct GPT-J | GPT-J | 6 | Inst |
| Meta | LlaMA 13B | LLaMA | 13 | Pre |
| Google | T5-11b-TQAO | T5 | 11 | QA |
| Bejing AI | GLM-10B | GLM | 10 | Pre |
| Google | T5-XXL-TQAO | T5 | 11 | QA |
| BigScience | BloomZ-7B | Bloom | 7 | Inst |
| Stanford / chavinlo | Alpaca native 7B | LLaMA | 7 | Inst |
| Bejing AI | GLM-130B | GLM | 130 | Pre |
| Meta | LlaMA 30B | LLaMA | 30 | Pre |
| Technology Innovation Institute, UAE | Falcon 40B | Falcon | 40 | Pre |
| Meta | LLaMA 7b | LLaMA | 7 | Pre |
| Data Bricks | Dolly v1 6B | GPT J | 6 | Inst |
| VMware | OpenLLaMA 7B Instr. | OpenLLaMA | 7 | Inst |
| Data Bricks | Dolly v2 12B | Pythia | 12 | Inst |
| Aleph Alpha | Luminous Supreme | Luminous | 70 | Pre |
| VMware | OpenLLaMA 7B OpenInst | OpenLLaMA | 7 | Inst |
| Meta | OPT-66B | OPT | 66 | Pre |
| Pankaj Mathur / TheBloke | Orca Mini 13B GPTQ | LLaMA | 13 | Inst |
| TogetherComputer | GPT-JT 6B | GPT-J | 6 | Inst |
| BigScience | mT0 XXL | mT5 | 11 | Inst |
| Cerebras | Cerebras 13B | Cerebras | 13 | Inst |
| Eleuther AI | Pythia 12B | Pythia | 12 | Pre |
| Meta | OPT-6.7b | OPT | 6.7 | Pre |
| Meta | OPT-IML 30B | OPT | 30 | Inst |
| Eleuther AI | GPT-J-6B | GPT | 6 | Pre |
| Eleuther AI | GPT Neo X 20B | GPT | 20 | Pre |
| Google | T5 v1.1 XXL | T5 | 11 | Pre |
| Meta | Galactica 30B | OPT | 30 | Inst |
| Meta | OPT-30B | OPT | 30 | Pre |
| Stability AI | StableLM-7b | StableLM | 7 | Pre |
| Meta | OPT-13B | OPT | 13 | Pre |

Appendix B:

**Authors**

Short Biography

Photo